\pdfoutput=1

\documentclass[11pt]{article}

\usepackage{acl}

\usepackage{times}
\usepackage{latexsym}
\usepackage{amsmath} 
\usepackage{booktabs} 
\usepackage{multirow}

\usepackage[T1]{fontenc}

\usepackage[utf8]{inputenc}

\usepackage{microtype}

\usepackage{inconsolata}

\usepackage{graphicx}

\usepackage{tcolorbox}
\tcbuselibrary{breakable} 

\title{Look Again, Think Slowly: Enhancing Visual Reflection in Vision-Language Models}

\author{
    Pu Jian\textsuperscript{1,2}, \ 
    Junhong Wu\textsuperscript{1,2},\ 
    Wei Sun \textsuperscript{1,2}, \ 
    Chen Wang \textsuperscript{1,2}, \ 
    Shuo Ren\textsuperscript{1}, \
    Jiajun Zhang\textsuperscript{1,2,3}\thanks{\ \ Corresponding Author}   \\
    \textsuperscript{1}Institute of Automation, Chinese Academy of Sciences\\
    \textsuperscript{2}School of Artificial Intelligence, University of Chinese Academy of Sciences\\
    \textsuperscript{3}Wuhan AI Research\\
    \texttt{\{jianpu2023, wujunhong2021, sunwei2023, chenwang2020, shuo.ren\}@ia.ac.cn}\\
    \texttt{jjzhang@nlpr.ia.ac.cn} \\
}

\begin{document}
\maketitle
\begin{abstract}

Recent advances in text-only “slow-thinking” reasoning have prompted efforts to transfer this capability to vision-language models (VLMs), for training visual reasoning models (\textbf{VRMs}). However, such transfer faces critical challenges: Effective "slow thinking" in VRMs requires \textbf{visual reflection}, the ability to check the reasoning process based on visual information. Through quantitative analysis, we observe that current VRMs exhibit limited visual reflection, as their attention to visual information diminishes rapidly with longer generated responses. To address this challenge, we propose a new VRM \textbf{Reflection-V}\footnote{The related codes are released in this URL: https://github.com/jian0805/ReflectionV}, which enhances visual reflection based on reasoning data construction for cold-start and reward design for reinforcement learning (RL). Firstly, we construct vision-centered reasoning data by leveraging an agent that interacts between VLMs and reasoning LLMs, enabling cold-start learning of visual reflection patterns. Secondly, a visual attention based reward model is employed during RL to encourage reasoning based on visual information. Therefore, \textbf{Reflection-V} demonstrates significant improvements across multiple visual reasoning benchmarks. Furthermore, \textbf{Reflection-V} maintains a stronger and more consistent reliance on visual information during visual reasoning, indicating effective enhancement in visual reflection capabilities.
\end{abstract}

\section{Introduction}

\begin{figure}[t]
  \centering
  \includegraphics[width=0.48\textwidth]{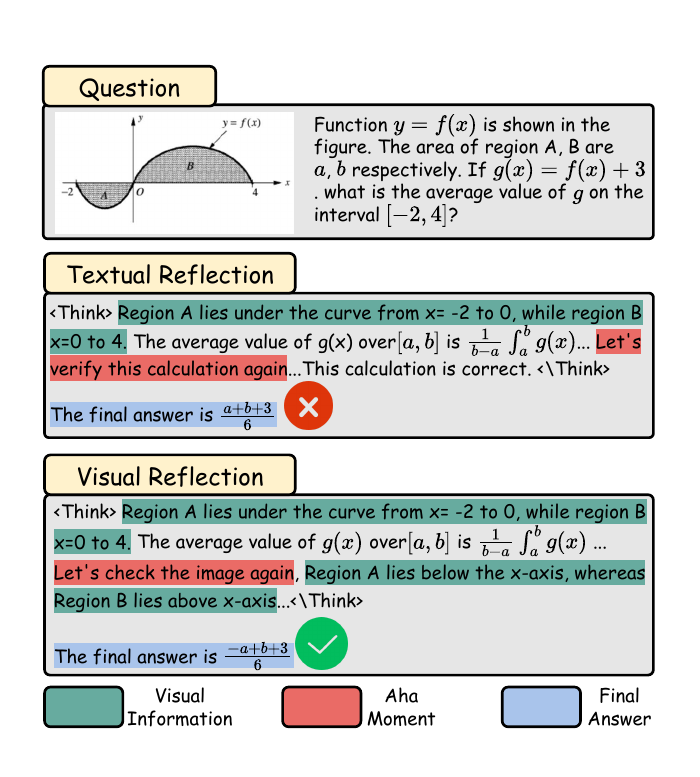}
  \caption{Existing “slow-thinking” VLMs claimed “aha moment” is often merely a textual reflection. We instead highlight visual reflection, where VLM actively verifies and refines its reasoning based on visual inputs.
  }
  \label{fig:intro}
\end{figure}

Recently, "slow-thinking" reasoning has emerged as a significant advancement in large language models (LLM) domain \cite{shao2024deepseekmath, sun2025ktae, chen2025ace}, demonstrating remarkable capabilities in solving complex reasoning tasks \cite{chen2025lr, xu2025hit}, such as OpenAI-o1 \cite{jaech2024openai} and DeepSeek-R1 \cite{guo2025deepseek}. The superior performance of "slow thinking" LLM primarily benefits from its ability to perform "reflection" during reasoning \cite{yan2024mirror}. This reflection mechanism allows models to check and revise intermediate steps before generating the final answer, thereby avoiding errors that may arise from short-cut inference \cite{snell2024scaling, yang2025towards, cheng2024vision}, which is also called “aha moment”. Inspired by this success, some researchers attempt to integrate "slow thinking" into vision-language models (VLMs), enabling the trained visual reasoning models (VRMs) to generate more accurate and deliberate solutions \cite{wang2025vl, chen2025sft, tan2025reason, huang2025vision}. 
Specifically, they leverage "slow thinking" LLMs to reason based on image descriptions generated by VLMs, thereby introducing reflection patterns into reasoning data. These visual reasoning data are often used for supervised fine-tuning (SFT) \cite{thawakar2025llamav, xu2024llava}, providing a cold-start initialization for subsequent reinforcement learning (RL) \cite{huang2025vision, tan2025reason}.

In this paper, we propose that the true “\emph{aha moment}” in visual reasoning arises when a model engages in \textbf{visual reflection}—that is, when it actively verifies and refines its reasoning based on the visual input, as shown in Figure~\ref{fig:intro}.  
However, current distillation-based approaches to training VRMs often miss this crucial aspect. By transferring superficial reflective behaviors from LLMs trained solely on text, these methods encourage reasoning patterns that are detached from the visual modality. This is because the cold-start data for these VRMs still originates from text-only reasoning on visual descriptions, and the RL stage uses rewards based solely on textual outputs \cite{huang2025vision, meng2025mm}. Thus, instead of promoting visual insight, these VRMs risk reinforcing textual hallucinations and visual neglect~\cite{zhong2024investigating, favero2024multi}. As a result, VRMs may appear reflective while actually bypassing the visual content, undermining both the reliability and robustness of their reasoning.

Consistent with the previous discussion, we conducted a detailed analysis of existing VRMs and found that they struggle with visual reflection. Specifically, experiments observe that existing VRMs' attention to and reliance on visual information decline rapidly as the number of generated tokens increases. And VRMs trained by distilling text-only reflection data even exhibit lower reliance on visual prompts than their backbone VLMs. This indicates that existing VRMs struggle to attend to and leverage visual information during reflection, thereby degrading into text-only reflection models.

To address the aforementioned challenges, we propose a novel two-stage training strategy for training VRMs. In the cold-start stage, we focus on resolving the limitations of image description-based approaches in incorporating visual reflection patterns within training data. Specifically, we leverage a multi-modal agent, where LLMs interact with VLMs, to complete reasoning in an LLM-VLM interleaved way. This data construction paradigm ensures that visual information can be continuously accessed and repeatedly utilized during reasoning, thereby introducing a visual reflection pattern for VRMs to learn. In the RL stage, to further promote the visual reflection behavior learned from cold-start data, we introduce a visual attention based reward for group relative policy optimization (GRPO) \cite{shao2024deepseekmath}. This reward encourages VRMs to consistently attend to visual information.

Reflection-V, our VRM trained with the proposed strategy, achieves significant improvements on benchmarks focusing on mathematical \cite{lu2023mathvista, wang2024measuring}, multi-disciplinary \cite{yue2024mmmu, yue2024mmmup}, and general reasoning \cite{chen2024m3cot}. At the 7B scale, it is comparable to or even surpasses several widely used very large VLMs, like GPT-4o \cite{hurst2024gpt} and InternVL2.5-38B \cite{chen2024expanding}. Notably, the aforementioned quantitative analyses and case study further show that compared to the base model, Reflection-V maintains more sustained attention to visual information and actively engages in visual reflection, representing the emergence of the true "\textit{aha moment}" in visual reasoning.

\section{VRMs Struggle with Visual Reflection}
\label{pre_exp}

In this section, we claim that existing VRMs struggle to perform visual reflection. To support this claim, we analyze the visual tokens' role during reasoning. Specifically, we quantify visual tokens' effect using the following metrics: attention weight and a visual dependency measure. This observation later motivates our proposed methodology.

\subsection{Visual Attention Weight}
\label{vis_attn}
To capture how the contribution of visual tokens varies during reasoning, we track the attention weights from response tokens to visual tokens as more tokens are generated. Let $T_{\text{res}}$  and $T_{\text{vis}}$ denote the sets of response and visual tokens, respectively.  For the $h$-th layer, let $a_{nj}^{(h)}$ represent the attention weight from the $n$-th response token to the $j$-th visual token. Thus the total attention from the $n$-th response token to $T_{\text{vis}}$ is given by

\begin{equation}
\text{Attn} (n, T_{\text{vis}}) = \frac{\sum_h \sum_{j \in T_{\text{vis}}} a^{(h)}_{nj}}{\sum_h\sum_{j \in T_{\text{vis}}} \mathbf{1}_{a^{(h)}_{nj} > 0}}.
\end{equation}

\subsection{Visual Dependency Measure}
Furthermore, after generating several tokens, we drop the visual tokens and assess VRMs' reliance on the visual token during reasoning by measuring the divergence in subsequent generations, which is quantified based on the divergence between the next-token prediction distributions with and without visual tokens. We use the Hellinger distance \cite{favero2024multi}, defined as 
\begin{equation}
H_{\text{dist}}(p, q) = 2^{-\frac{1}{2}} \sqrt{\sum_{i=1}^{k} \left( \sqrt{p_i} - \sqrt{q_i} \right)^2}
\end{equation}
to quantify the divergence between two probability distributions. Therefore, when the number of generated response tokens is $n$, the visual dependency measure $\text{VDM}(n|T_{\text{vis}}, T_{q})$ for a given image-question pair $(T_{\text{vis}}, T_{q})$ is given by 
\begin{equation}
H_{\text{dist}} \left( p(\cdot | T^{<n}_{res}, T_{\text{q}}, T_{\text{vis}}), p(\cdot | T^{<n}_{res}, T_{\text{q}}) \right).
\end{equation}

\subsection{Diminishing Visual Attention and Dependency}

\begin{figure}[t]
  \centering
  \includegraphics[width=0.47\textwidth]{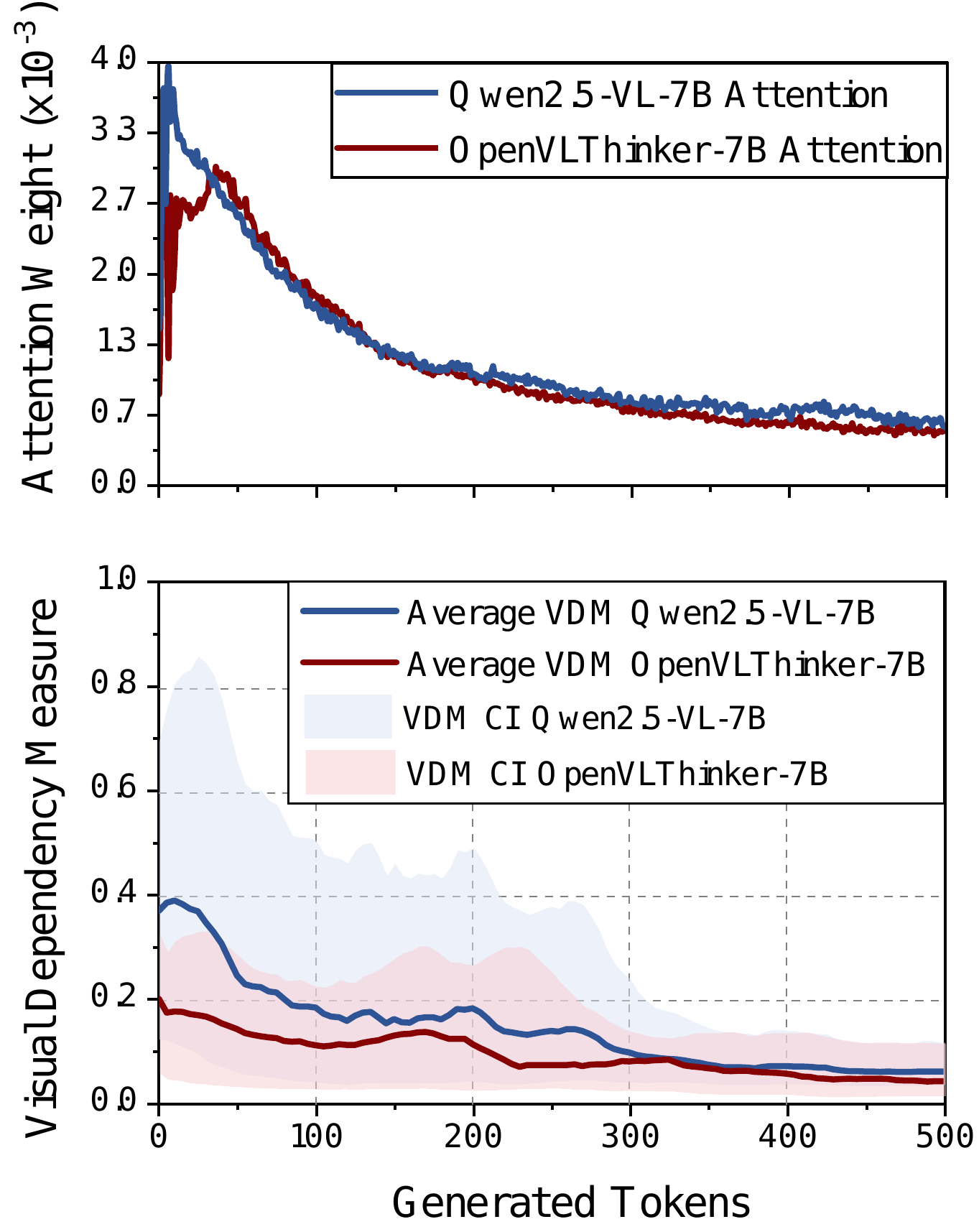}
  \caption{Attention weights on visual tokens and the visual dependency measure during reasoning on the MMMU dataset. Both metrics decline sharply as more tokens are generated, and RL-enhanced models (e.g., OpenVLThinker-7B) do not mitigate this decay.
  }
  \label{fig:vdm_attn}
\end{figure}

As Figure \ref{fig:vdm_attn} illustrates, both the mean Visual Dependency Measure on MMMU \cite{yue2024mmmu} and the layer-wise attention from response tokens to visual tokens in VRM, decline sharply as generation proceeds: after roughly 300 tokens, visual attention falls to only 20–30 \% of its initial level. This analysis reveals that VRMs typically lack visual reflection ability in long-chain reasoning, as they rarely refer back to visual tokens when performing reflective checking of reasoning process. 

Although widely used RL boosts VRMs’ reasoning performance, it fails to equip them with visual reflection capability. Instead, it further reinforces over-reliance on previously generated text. As Figure~\ref{fig:vdm_attn} shows, OpenVLThinker, RL based on Qwen2.5-VL, exhibits even lower focus and dependence on visual tokens during reasoning than the base model.

\section{Method}
\label{method}

In the above analysis, we observe that as the length of reasoning process increases, VRMs rapidly reduce their reliance on and attention to visual information. This limitation hampers their ability to perform visual reflection and prevents them from fully benefiting from "slow thinking" paradigms like DeepSeek-R1 \cite{guo2025deepseek}. To address this problem, we propose a two-stage strategy for training VRMs. This strategy consists of: (1) cold-start initialization \cite{yang2025r1} on reasoning data with visual reflection, and (2) reinforcement learning with a visual attention-based reward. In the first stage, we leverage a multi-modal agent, where LLMs interact with VLMs, to construct visual reasoning data exhibiting visual reflection, and use it to perform supervised fine-tuning (SFT) on the base VLM. The second stage applies GRPO with the proposed reward function that explicitly encourages sustained attention to visual tokens. The related codes can be found at https://github.com/jian0805/ClearVQA

\subsection{Reasoning Data with Visual Reflection Construction}
\begin{figure*}[t]
  \centering
  \includegraphics[width=0.995\textwidth]{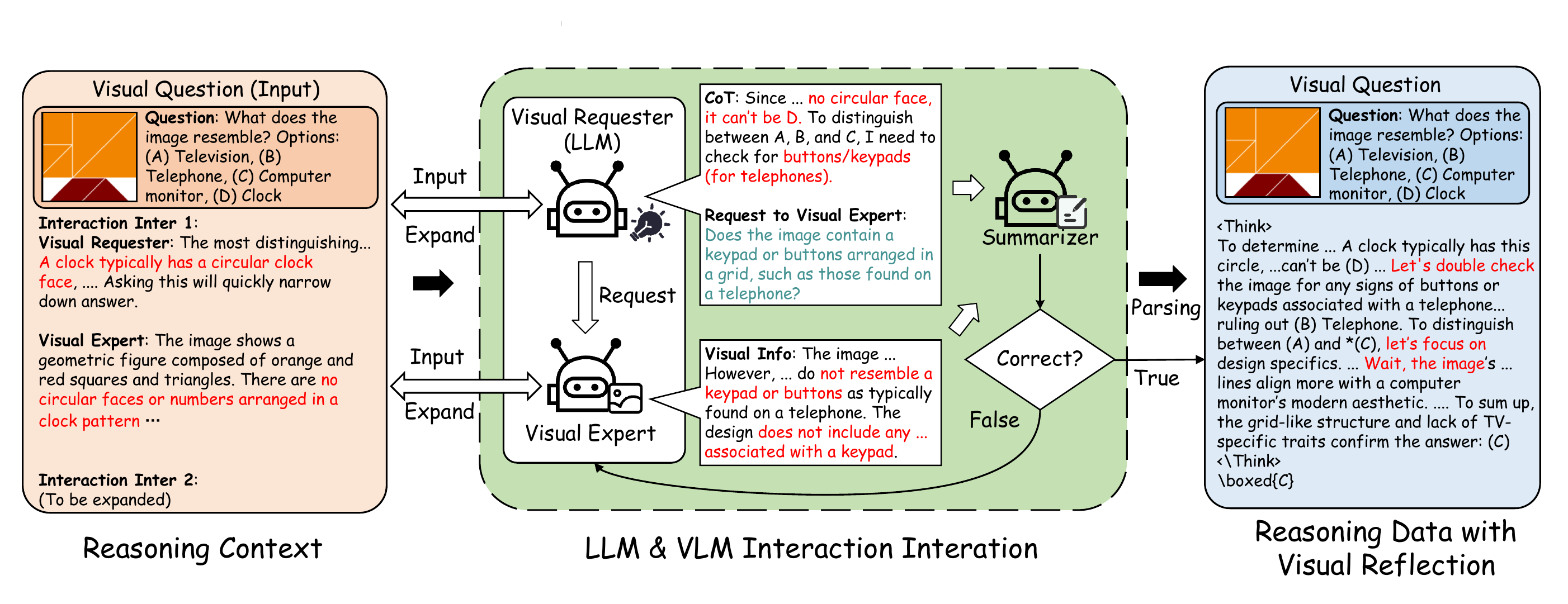}
  \caption{A workflow for constructing reasoning data with a visual reflection pattern. LLMs and VLMs perform reasoning through interaction, ensuring that visual information is continuously acquired and repeatedly utilized, thereby creating a visual reflection pattern in the reasoning process.
  }
  \label{fig:data_method}
\end{figure*}

Existing visual reasoning studies typically rely on LLMs to perform reasoning directly based on image captions, thereby constructing visual reasoning data \cite{liu2024improved, yu2024capsfusion, huang2025vision}. The absence of visual modality during reasoning makes it impossible for these reasoning data to exhibit the visual reflection we claimed. Inspired by recent advances in visual-language agents \cite{zhou2024proreason, jian2024large}, we employ crafted prompts to guide the interaction between the LLM and VLM in completing the reasoning task. This data construction paradigm ensures that visual information is continuously accessible and repeatedly utilized during reasoning, thereby introducing a visual reflection pattern. The data construction process is detailed in the following sections.

\textbf{Visual Reflection Data Construction Based on LLM-VLM Interaction}. As shown in Figure~\ref{fig:data_method}, in our data construction process, the LLM and VLM interact by taking on the following three roles: 1) \textbf{Visual requester}, played by the LLM, which determines what visual information is needed to answer the question based on the existing reasoning context and proposes a request to the VLM. During this process, the CoT output from the VLM is added as a partial solution to the reasoning context. 2) \textbf{Visual responder}, played by the VLM, replies to the request made by the visual requester, revealing visual information related to the visual question. The visual description generated by the visual responder is added to the reasoning context. 3) \textbf{Summarizer}, played by the LLM, summarizes the existing reasoning context after each round of interaction, generating the final answer. If the generated answer does not match the ground truth, all outputs from the summarizer are discarded, and a new round of interaction begins. Otherwise, the output is added to the reasoning context, completing the reasoning process for the visual question. 

\textbf{Post-generation processing}. After constructing visual reflection data through LLM and VLM interactions, we perform the following steps to ensure data quality: 1) \textbf{Non-Reflection Filtering}. We filter out the data where the summarizer produces the correct answer just after the first interaction. These samples lack sustained reliance on visual information, and the reasoning solution does not exhibit a visual reflection pattern. 2) \textbf{Cohesion Enhancement}. In the previous data construction process, the text generated across different VLM-LLM interaction rounds may lack coherence. We employ the LLM to process and refine the text into a cohesive reasoning process. All detailed prompts are provided in Appendix~\ref{apppenidx:prompts}.

\subsection{Visual Attention Based Reward}
Following existing works \cite{wei2025skywork, xiao2025fast, zhang2025r1}, we adopt GRPO, a rule-based reinforcement learning algorithm, to enhance the reasoning capabilities of VRMs. Building on the original reward function, we introduce a visual attention–based reward to encourage the model to maintain sustained attention to visual tokens throughout the reasoning process while preserving overall performance.

Specifically, based on the analyses in Section~\ref{vis_attn} that attention weights on visual tokens rapidly decrease as the number of generated tokens increases, our reward rule follows the principle: For a VRM reasoning process, VRMs receive a higher reward if relatively higher attention to visual tokens is maintained after generating several tokens. Therefore, the visual attention based reward is given by

\begin{equation}
r_{v} = \begin{cases}
    \frac{\sum\limits_{n>|T_{res}|/2} \text{Attn} (n, T_{\text{vis}})}{\sum\limits_{n<|T_{res}|/2} \text{Attn} (n, T_{\text{vis}})} & \text{if } r_a = 1 \\
    0 & \text{if }  r_a = 0
\end{cases} .
\end{equation}

\begin{table*}[tb]
\small
\centering
\setlength{\tabcolsep}{4pt}  
\renewcommand{\arraystretch}{1.25} 
\begin{tabular}{lcccccc}
\toprule
\textbf{Model} & \multicolumn{2}{c}{\textbf{Math-Reasoning}} & \multicolumn{2}{c}{\textbf{Multi-Disciplinary}} & \multicolumn{1}{c}{\textbf{General}} &  \multicolumn{1}{c}{\textbf{Hallucination}}  \\
\cmidrule(lr){2-3} \cmidrule(lr){4-5} \cmidrule(lr){6-6} \cmidrule(lr){7-7}
 & MathVision & MathVista & MMMU & MMMU-Pro & M3CoT & HallBench\\
\midrule
\multicolumn{7}{c}{\textbf{Closed-Source Vision-Language Models}} \\
\midrule
GPT-4o \cite{hurst2024gpt} & 30.4 & 60.0 & 69.1 & 51.9 & $^\dagger$74.2 & - \\
GPT-4V \cite{yang2023dawn} & 22.8 & 49.9 & 56.8 & 41.2 & 62.6 & \textbf{65.3} \\
\midrule
\multicolumn{7}{c}{\textbf{Open-Source Vision-Language Models}} \\
\midrule
QwenVL2.5-3B \cite{bai2025qwen2} & 21.2 & 62.3 & $^\dagger$51.2 & 31.6 & $^\dagger$55.6 & 45.1\\
QwenVL2.5-7B \cite{bai2025qwen2} & 25.1 & 68.2 & $^\dagger$54.3 & 36.9 & $^\dagger$60.5 & 49.5\\
InternVL2.5-8B \cite{chen2024expanding} & 19.7 & 63.6 & 56.0 & 30.5 & $^\dagger$41.5 & 49.0 \\
InternVL-2.5-38B \cite{chen2024expanding} & \underline{32.2} & 71.9 & \underline{57.6} & \textbf{46.0} & $^\dagger$\underline{68.9}  & - \\
LLaVA-OneVision-72B \cite{li2024llava}  & 30.1 & 67.5 & 56.8 & 31.0 & $^\dagger$61.5 & 47.9\\
Kimi-VL-16B \cite{team2025kimi} & 21.4 & 68.7 & 55.7 & - & - &- \\
\midrule
\multicolumn{7}{c}{\textbf{Open-Source Vision-Language Reasoning Models}} \\
\midrule
TVC-7B \cite{sun2025mitigating} & 22.7 & 68.1 & - & - & - & - \\
R1-VL-7B \cite{zhang2025r1} & 24.7 & 63.5 & 44.5 & - & - & -\\
MM-Eureka-7B$^\diamond$ \cite{meng2025mm} & 26.9 & \underline{73.0} & $^\dagger$51.3 & $^\dagger$36.7 & $^\dagger$63.5 & 47.8 \\ 
R1-Onevision-7B$^\diamond$ \cite{yang2025r1} & 29.9 & 64.1 & $^\dagger$48.7 & 21.6 & $^\dagger$53.1 & 41.7 \\
OpenVLThinker-7B$^\diamond$ \cite{deng2025openvlthinker} & 25.3 & 70.2 & 52.5 & 37.3 & $^\dagger$62.2 & 42.3 \\
\midrule
\multicolumn{7}{c}{\textbf{Ours} (\textit{Training strategy emphasizing visual reflection})} \\
\midrule
\textbf{Reflection-V-3B} & 27.9 & 66.3 & 56.9 & 38.2 & 62.9  & 49.3 \\
\textbf{Reflection-V-7B} & \textbf{33.9} & \textbf{73.3} & \textbf{61.3} & \underline{42.7} & \textbf{71.1} & \underline{53.9}\\
\bottomrule
\end{tabular}
\caption{\small Performance of Reflection-V across various visual reasoning benchmarks, compared to existing VLMs. $\dagger$ indicates results reproduced by us. $\diamond$  denotes vision-language reasoning Models also based on the Qwen2.5-7B series. Bold and underlined scores represent the best and second-best performance among open-source models for each benchmark.}
\label{table:main_results}
\end{table*}

\begin{table*}[tb]
\small
\centering
\renewcommand{\arraystretch}{1.2} 
\begin{tabular}{lccccc}
\toprule
Model & MathVision & MathVista & MMMU & MMMU-Pro & M3CoT \\ 
\midrule
\textbf{Reflection-V-3B} & 27.94 & 66.30 & 56.89 & 38.17 & 62.95 \\
\textit{~~~w/o VAR} & 26.52 & 65.60 & 55.80 & 36.56 & 61.79 \\
\textit{~~~w/o Cold-Start} & 24.27 & 64.20 & 53.98 & 34.75 & 59.55 \\
\textit{~~~w/o Cold-Start + VAR} & 23.60 & 63.90 & 53.21 & 33.97 & 58.81 \\
\midrule
\textbf{Reflection-V-7B} & 33.71 & 73.30 & 61.33 & 42.71 & 71.07 \\
\textit{~~~w/o VAR} & 32.47 & 72.40 & 60.10 & 41.95 & 69.28 \\
\textit{~~~w/o Cold-Start} & 29.01 & 70.40 & 58.81 & 39.06 & 65.87 \\
\textit{~~~w/o Cold-Start + VAR} & 28.53 & 69.80 & 58.03 & 38.24 & 64.63 \\
\bottomrule
\end{tabular}
\caption{Ablation results for cold-start based on visual reflection data and visual attention based reward on performance improvement. VAR denotes visual attention based reward.}
\label{table:ablation}
\end{table*}

\begin{table*}[tb]
\small
\centering
\renewcommand{\arraystretch}{1.2} 
\begin{tabular}{lccccc}
\toprule
Model & MathVision & MathVista & MMMU & MMMU-Pro & M3CoT \\
\midrule
\textbf{Reflection-V-3B} & 27.96 & 66.30 & 56.89 & 38.17 & 62.95 \\
VR SFT $\rightarrow$ Cap\&R SFT & 25.04 & 63.90 & 54.22 & 33.59 & 60.41 \\
\midrule
\textbf{Reflection-V-7B} & 33.88 & 73.30 & 61.33 & 42.71 & 71.07 \\
VR SFT $\rightarrow$ Cap\&R SFT & 29.31 & 69.00 & 58.41 & 37.95 & 66.25 \\
\bottomrule
\end{tabular}
\caption{Comparative results of cold-start initialization using data with visual reflection pattern and mage caption-based reasoning data on visual reasoning performance improvement.}
\label{table:ablation_cap_r_cold_start}
\end{table*}

Here, $r_a$ is the accuracy reward taking values from \{0, 1\}. Refer to function (2), $ \text{Attn} (n, T_{\text{vis}})$ represents the average attention weight of the $n$-th response token to the visual tokens (averaged over all attention heads). $|T_{res}|$ denotes the total number of tokens in a VRM's reasoning process.

Based on the observation shown in Figure~\ref{fig:vdm_attn}, we calculate the visual attention based reward using the last layer where the attention to visual tokens is most significant. The overall reward $r_{o}$ in GRPO is the weighted sum of the accuracy reward $r_a$, visual attention-based reward $r_v$, and format reward $r_f $ \cite{shao2024deepseekmath}, given by
\begin{equation}
r_{o} = r_a + \lambda_v r_v + \lambda_f r_f .
\end{equation}
$\lambda_v$ and $\lambda_f$ are scaling coefficients set to 0.5 and 0.1, respectively.

\section{Experiments}

\subsection{Experimental Setup}
\textbf{Implementations}. To construct the cold-start data, we use the open-source VLM Qwen-2.5-VL-72B \cite{bai2025qwen2} and a reasoning-capable LLM, QWQ-32B \cite{team2025qwq}, to interactively generate data. Our method is evaluated using the Qwen-2.5-VL-7B-Instruct as the base model. During the cold-start stage, we train for 3 epochs on 2 NVIDIA H100 GPUs. The model, after cold-start initialization, is subsequently trained using GRPO with visual attention based reward for 12 epochs on 8 NVIDIA H100 GPUs, based on the Verl training framework \cite{sheng2024hybridflow, zheng2025easyr1}. For GRPO, 16K reasoning samples are collected from a diverse multimodal corpus. The detailed composition of training data is shown in Appendix~\ref{apppenidx:data_statistic}. Train details
for cold-start initialization and GRPO stages is provided in Appendix~\ref{apppenidx:hyperparameter}.

\textbf{Benchmarks for Evaluation}. We conduct a comprehensive experimental analysis to assess how our method improves visual reasoning. To ensure a well-rounded evaluation, we select widely recognized visual reasoning benchmarks that emphasize math, multi-disciplinary, and general reasoning skills. For evaluating math reasoning, we use MathVista \cite{lu2023mathvista} and MathVision \cite{wang2024measuring}, which are standard tests for visual reasoning models. To evaluate performance across multiple disciplinary such as physics, chemistry, and computer science, we adopt MMMU and MMMU-Pro \cite{yue2024mmmu, yue2024mmmup}. Furthermore, M3CoT \cite{chen2024m3cot} is used to assess general reasoning ability, as it covers a broad range of knowledge-intensive and commonsense-based reasoning questions.

\subsection{Main Result}

\begin{table*}[tb]
\small
\centering
\renewcommand{\arraystretch}{1.3} 
\begin{tabular}{lccccc}
\toprule
Model & MathVision & MathVista & MMMU & MMMU-Pro & M3CoT \\
\midrule
Reflection-V-3B (\emph{Qwen2.5-VL/QWQ cold start}) & 27.9 & 66.3 & 56.9 & 38.2 & 62.9 \\
Reflection-V-3B (\emph{InternVL3/Qwen3 cold start}) & 27.1 & 67.6 & 58.0 & 36.4 & 64.2 \\
\bottomrule
\end{tabular}
\caption{Performance comparison between cold-start data constructed with InternVL3/Qwen3 and Qwen2.5-VL/QWQ. In this experiment, InternVL3-38B, Qwen3-32B, and Qwen2.5-VL-72B are employed.}
\label{table:model_family_ab}
\end{table*}

\begin{figure*}[t]
  \centering
  \includegraphics[width=0.99\textwidth]{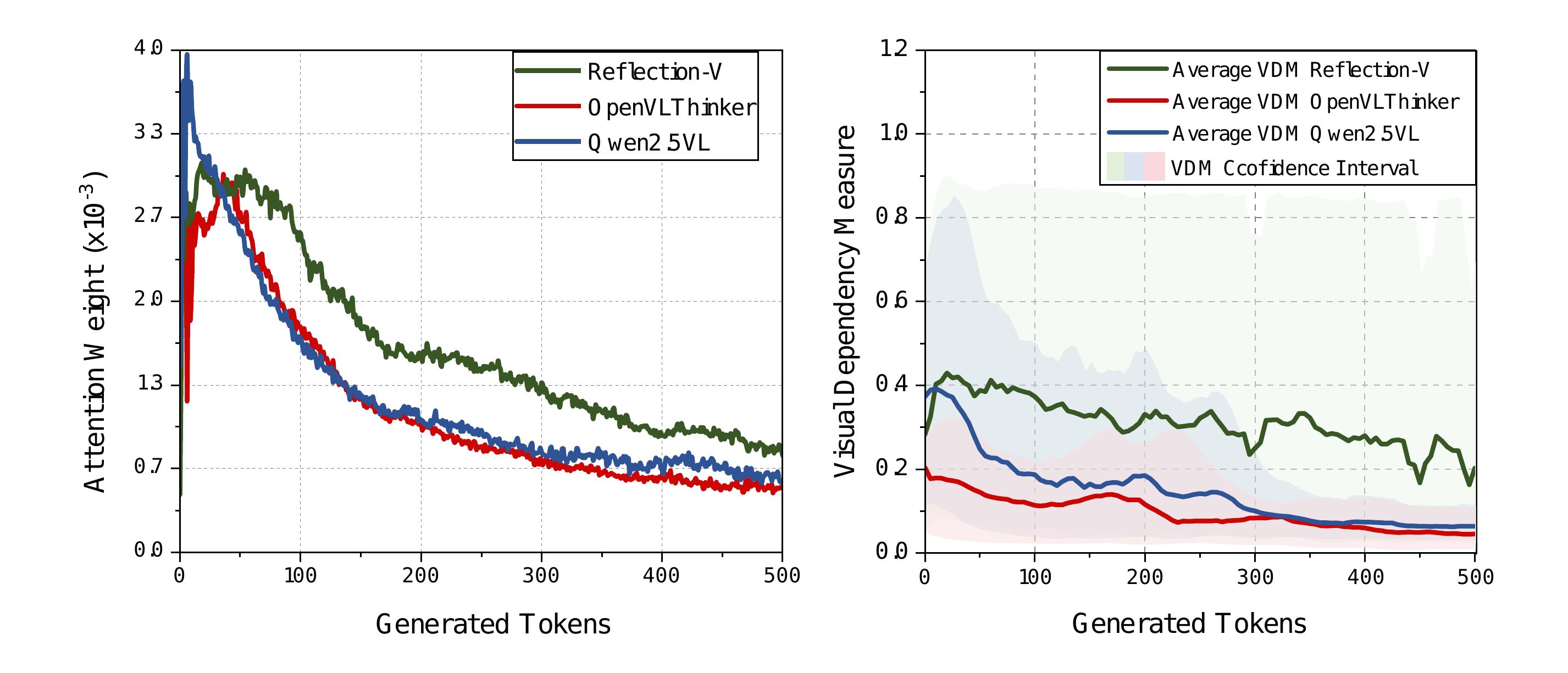}
  \caption{Attention weights (last layer) on visual tokens and visual dependency measure of Reflection-V-7B on MMMU benchmark, compared to OpenVLThinker-7B and Qwen2.5VL-7B. The shown attention weights represent the mean value across all samples. Visual dependency measure quantifies the difference in probability distributions for next token prediction based on generated tokens, before and after discarding visual tokens. The light-green, light-blue, and light-red bands represent the confidence intervals of the visual dependency measure for Reflection-V-7B, Qwen 2.5-VL-7B, and OpenVLThinker-7B, respectively.
  }
  \label{fig:atten_weight_reflection}
\end{figure*}

We evaluate the performance of our model, Reflection-V, on visual reasoning benchmarks across three categories: math, multi-disciplinary, and general, as shown in Table~\ref{table:main_results}. The results indicate that our model significantly outperforms Qwen2.5-VL \cite{bai2025qwen2} base model and other open-source models of similar scale in reasoning capability. Even compared to existing vision-language reasoning models based on RL, Reflection-V-7B achieves a notable margin of improvement.

Notably, Reflection-V-7B reaches comparable or even superior performance compared to some widely used, large-scale closed-source and open-source VLMs. For instance, on MathVision and MathVista, Reflection-V-7B outperforms GPT-4o and InternVL-2.5-38B \cite{chen2024expanding}. On MMMU and M3CoT, Reflection-V-7B surpasses InternVL-2.5-38B and LLaVA-OneVision-72B \cite{li2024llava}, and is comparable to GPT-4o \cite{hurst2024gpt}. On MMMU-Pro, Reflection-V-7B outperforms LLaVA-OneVision-72B and GPT-4V \cite{yang2023dawn}, while being comparable to InternVL-2.5-38B. In contrast to existing vision-language reasoning models, which show improved math reasoning but a decline in multi-disciplinary and general reasoning capabilities, Reflection-V demonstrates improvements across all three categories. Additionally, experimental results show that the proposed method is effective across models of different scales.

Surprisingly, thanks to "visual reflection" we proposed, Reflection-V exhibits significantly fewer visual hallucinations. Specifically, we use HallBench \cite{guan2024hallusionbench} to quantify the extent of visual hallucinations in VLMs or VRMs. Compared to the Qwen2.5-VL base model, Reflection-V shows a significantly higher all accuracy (aAcc) \cite{guan2024hallusionbench} on HallBench. This result suggests that, due to sustained focus on visual information, the visual hallucinations commonly observed in VRMs are notably suppressed. In contrast, other VRMs experience even more severe visual hallucinations than the base model. This can be attributed to the fact that, as mentioned earlier, VRMs struggle to maintain sustained attention to visual information as more textual tokens are generated during reasoning.

\subsection{Ablation Study}

\begin{figure*}[t]
  \centering
  \includegraphics[width=0.99\textwidth]{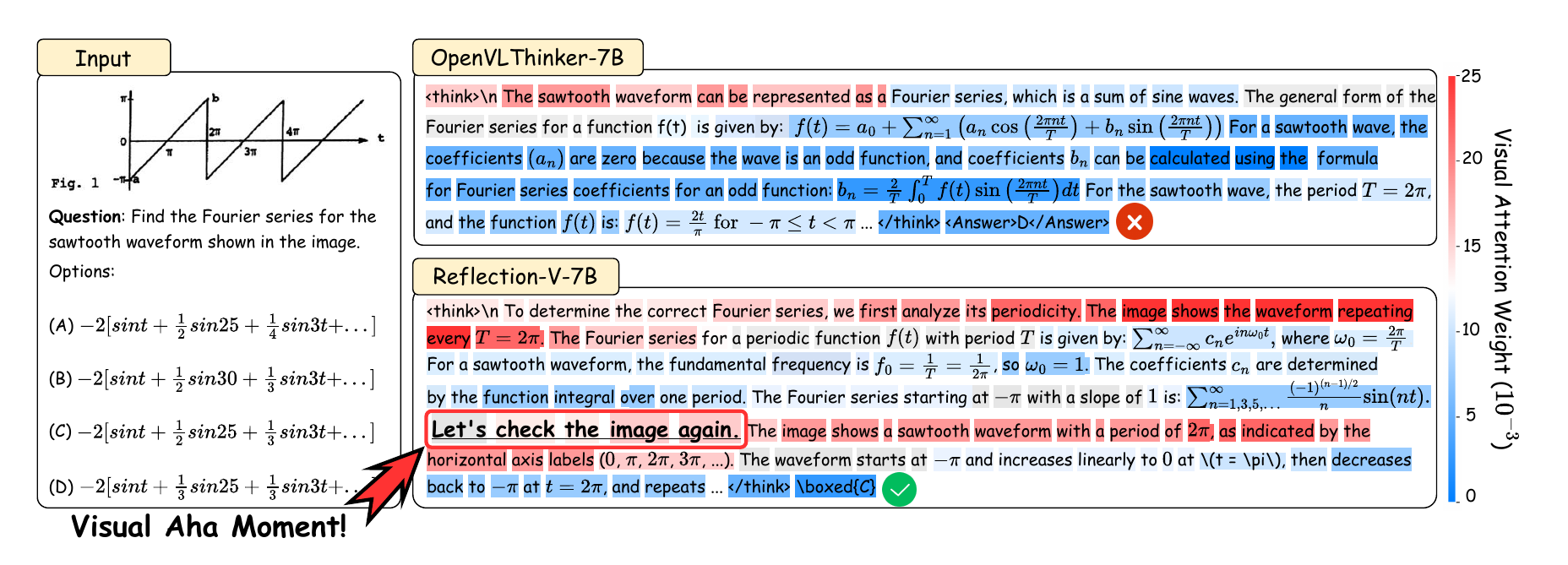}
  \caption{Our model, Reflection-V, exhibits the visual reflection capacity that we claim, in contrast to other RL-based visual reasoning models. The background color of tokens in the figure indicates the magnitude of the visual attention weight. This visual reflection capacity is demonstrated by the recheck and attention to visual tokens again that appear alongside "Aha moment", like "Let's check the image again".
  }
  \label{fig:case_study}
\end{figure*}

We ablate the cold-start and visual attention based reward components of our method to evaluate the impact of each design on enhancing visual reasoning capabilities of VRMs. Experimental results in Table~\ref{table:ablation} demonstrate that both components significantly improve VRM's performance. Notably, cold-start yields a particularly substantial performance gain. This indicates that emphasizing the continuous reliance and repeated utilization of visual information in SFT data significantly improves visual reasoning performance. Furthermore, with cold-start, the performance improvement from visual attention based reward becomes more pronounced. We believe this occurs because cold-start, based on visual reflection data, guides VRMs in how to increase their attention to visual information. 

We conduct a further ablation to validate the superiority of emphasizing visual reflection pattern in cold-start data. Specifically, we replace the reasoning data containing visual reflection patterns with image caption-based reasoning data (derived from the same origin data) during cold-start initialization, then compare their performance across benchmarks. As shown in Table~\ref{table:ablation_cap_r_cold_start}, our method outperforms the "caption then reasoning" data construction paradigm by a significant margin. This result also illustrates that the improved visual reasoning performance originates not from distilling high-quality data from larger models but rather from the intentional incorporation of visual reflection patterns in constructed data.

Besides, We conducted additional experiments to analyze whether the proposed cold-start data construction method exhibits any bias toward specific model families. Specifically, we replaced the VLM and LLM used for constructing cold-start data with InternVL3-38B \cite{zhu2025internvl3} and Qwen3-32B (Thinking mode) \cite{yang2025qwen3}, and compared the results with those from constructing cold-start data using Qwen2.5VL-72B and QWQ-32B. The results are shown in Table~\ref{table:model_family_ab}. The experimental findings indicate that the performance gap between InternVL3/Qwen3 and Qwen2.5-VL/QWQ in constructing cold-start data is minimal. This suggests that our cold-start data construction method does not exhibit any bias toward specific model families.

To further validate that the proposed method improves performance by achieving the claimed visual reflection, we present further analyses below.

\subsection{Analyses}

In Section~\ref{pre_exp}, we demonstrate that existing VRMs struggle with visual reflection through three metrics: visual attention weight, and visual dependency measure. Based on these metrics, in this subsection, we analyze whether the performance improvement of Reflection-V genuinely stems from the training strategy that emphasizes visual reflection.

\textbf{Our method leads to more sustained visual attention and dependence}. We compare the attention weight of response tokens to visual tokens for Reflection-V and OpenVLThinker-7B of the same scale, at different generated tokens. We find that, in the middle and deep transformer layers, Reflection-V exhibits significantly higher attention weights to visual tokens than OpenVLThinker, the model also trained through SFT cold-start initialization and GRPO, based on Qwen2.5-VL, as shown in Figure~\ref{fig:atten_weight_reflection}. As the number of generated tokens increases, the decrease in attention weight to visual tokens is slower in Reflection-V-7B than in OpenVLThinker-7B. Furthermore, to investigate whether the proposed method enhances VRMs' reliance on visual information, we compare the visual dependency measure, as referred to in equation (3), of Reflection-V and OpenVLThinker at different generated tokens, as shown in Figure~\ref{fig:atten_weight_reflection}. The results indicate that, benefiting from the emphasis on visual reflection, Reflection-V significantly mitigates the diminishing of dependence on visual information as generated tokens increase, compared to OpenVLThinker. To sum up, Reflection-V exhibits more sustained reliance on visual information. Experimental results also show that this feature enhances visual reasoning accuracy.

\textbf{Better performance, sustained visual attention, and reliance all originate from visual reflection}. As discussed earlier, the proposed method improves visual reasoning performance while maintaining visual attention and dependency during reasoning. Figure~\ref{fig:case_study} presents a comprehensive example demonstrating that these gains are indeed due to the model’s visual reflection ability. In this example, Reflection-V actively verifies and refines its reasoning by rechecking the visual input. When textual “Aha moments” like “Let’s check the image again” appear, the visual attention weight rises sharply during next-token prediction, representing the true “aha moment” in visual reasoning. As Figure~\ref{fig:case_study} shows, visual reflection capacity enables Reflection-V to reveal critical visual information absent from the reasoning context, thereby inferring the correct answer. Additionally, Figure~\ref{fig:atten_weight_reflection} shows that the upper bound of the confidence interval for visual dependency measure shows virtually no decline with increasing generated tokens. This suggests that during reasoning, as the number of generated tokens increases, Reflection-V maintains consistent dependency on visual tokens throughout the generation process. The observed decline in average visual dependency measure stems from the reduced frequency of visual reflection as the number of generated tokens increases. This phenomenon aligns with the re-emergent, image-focused attention derived from visual reflection, which is observed in Figure~\ref{fig:atten_weight_reflection}. These results demonstrate that, when Reflection-V engages in visual reflection, it maintains the same level of focus and reliance on visual tokens as at the start of reasoning.


\section{Related Works}

\textbf{Visual Reasoning model.} Large VLMs typically project inputs from non-text modalities into textual representations that LLMs can process \cite{guan2025trifine, ye2025coherent, guo2025crop, jing2024dq, zhang2025mulcogbench}, achieving strong performance in vision understanding \cite{bai2025qwen2, chen2024expanding, jian2024large, zhang2025mllms}. The advancement of LLMs has redefined state-of-the-art performance across a vast landscape of tasks \cite{zhang2025simple, zhang2025discovering, ren2025towards, xu2024bridging}. 
Motivated by recent advances in LLM domain \cite{chen2023chinesewebtext, zhang2024chinesewebtext, chen2025ladm}, researchers enhance Large VLMs reasoning with step-level reasoning SFT datasets \cite{xu2024llava, jian2025teaching} and RL \cite{yang2025r1}. However, as discussed earlier, these trained VRMs typically struggle with visual reflection, leading VRMs to reason without visual grounding after many tokens are generated.

\textbf{Visual forgetting alleviation}. Consistent with visual reflection that we claim, some recent studies emphasize alleviating forgetting visual cues during long-chain inference. M3ID \cite{favero2024multi} employs mutual information decoding to amplify image influence while weakening linguistic priors, thereby promoting continuous reliance on visual cues. But diminishing linguistic priors lowers performance on complex reasoning tasks \cite{bitton2024visual, zhang2024mathverse}. TVC \cite{sun2025mitigating}, a concurrent work, periodically replays visual tokens during inference to reuse visual cues, but it cannot flexibly invoke visual reflection when required. Distinct from these works, we embed visual reflection capability into VRMs based on data generated by LLM-VLM interaction, and reinforce this capability during RL. As a result, the trained VRMs can actively refine their reasoning based on the visual input when needed.

\section{Conclusion}
In this paper, we propose that the true “aha moment” in visual reasoning arises when a model engages in visual reflection—that is, when it actively verifies and refines its reasoning based on the visual input. Through quantitative studies, we reveal that existing VRMs struggle with such visual reflection. Therefore, to address this critical challenge, we propose a two-stage training strategy combining LLM-VLM interaction-driven reflective reasoning patterns with visual attention based RL. This training strategy significantly improves performance across multiple benchmarks. Experiments confirm that such improvement is derived from sustained visual attention and reliance, demonstrating the effectiveness of visual reflection. This work establishes a foundation for integrating visual reflection into VRMs, narrowing the gap between visual and text-only reasoning on complex tasks.

\section*{Limitations}
Firstly, due to computational constraints, we could not conduct experiments like GRPO with models larger than 7B parameters. Thus, we limited our exploration to the 3B and 7B parameter scales. Secondly, our cold-start initialization (based on constructed reasoning data with visual reflection pattern), reinforcement learning (with visual attention based reward), and evaluation presently involve relatively limited categories of visual-language datasets. In future work, we plan to include a wider range of visual-language datasets covering diverse problem types to further evaluate the generalization ability of the proposed method.

\section*{Acknowledgments}
We thank the Wuhan AI Research for providing the computational resources. We also appreciate the outstanding open-source code repositories contributed by the LLM and multimodal communities. Finally, we thank all the reviewers for their detailed reviews and insightful comments.


\bibliography{acl_latex}

\clearpage

\appendix
\section{Implementation Details and Hyperparameters}
\begin{table}[t]
\small
\centering
\renewcommand{\arraystretch}{1.2}
\begin{tabular}{c|c}
\toprule
Hyper-parameters & Value \\ \hline
Epoches & 3 \\
Batch size & 8 \\
Warmup ratio & 0.1 \\
Gradient accumulation & 4 \\
Learning rate scheduler & Cosine \\
GPUs & 2 \\
Optimizer & AdamW \\ \bottomrule
\end{tabular}
\caption{The hyperparameters used during cold-start initialization using the constructed data with visual reflection pattern.}
\label{tab:hp_cold_start}
\end{table}

\begin{table}[t]
\small
\centering
\renewcommand{\arraystretch}{1.2}
\begin{tabular}{c|c}
\toprule
Hyper-parameters & Value \\ \hline
Hyper-parameters & 12  \\
Batch size & 512 \\
Micro Batch size & 8 \\
Warmup & False \\
Rollout & 16 \\
Rollout Temperature& 1.0 \\
Rollout Top-P& 0.99 \\
Freeze Vision Tower & True \\
KL divergence coefficient & $1\times10^{-2}$ \\
Learning rate & $5\times10^{-6}$ \\
Weight Decay & $1\times10^{-2}$ \\
GPUs & 8 \\
Optimizer & AdamW \\ 
Framework & Verl \\ 
\bottomrule
\end{tabular}
\caption{The hyper-parameters used during GRPO with visual attention based reward.}
\label{tab:hp_grpo}
\end{table}

\label{apppenidx:hyperparameter}
During supervised fine-tuning (SFT), we set the learning rate to 1e$^{-5}$, apply a cosine scheduler with a 0.1 warm-up ratio, use BF16 mixed precision, adopt a batch size of 8, and train for 3 epochs.
In the reinforcement-learning (RL) phase, we limit both prompts and responses to 2048 tokens and apply KL divergence with a coefficient of 1e$^{-2}$. Each training step processes 512 questions with 16 rollouts per question; rollout sampling uses a temperature of 1.0 and a top-p of 0.99.
For validation, we evaluate with the pass@1 metric and set the sampling temperature to 0.5. Detailed hyperparameters are shown in Table~\ref{tab:hp_cold_start} and Table~\ref{tab:hp_cold_start}.

\section{Prompts}
\label{apppenidx:prompts}
\begin{tcolorbox}[
    breakable,
    title=Prompt Templates of Visual Requester,
    colback=white,              
    colframe=blue!50!black,      
    fonttitle=\bfseries\itshape,   
    arc=2pt,                      
    left=5mm,                      
    right=5mm,
    top=3mm,
    bottom=3mm
]
\texttt{You currently need to address the following question: <question> The information you need is in an image, but you can't see the image right now. At the same time, you're not capable of complex reasoning. \\
\\
However, you can can consult the following two Vision Expert for help. You can ask him a single question for information in the picture, for example, you could ask him, "What color is the bird in the picture?"\\
\\
Use the following format: \\
\{'Thought': 'analyze the problem here.', 
 'Question':'Questions you want to ask the Vision EXPERT'\}\\
 \\ 
<split>\\
And the information you know currently is as follows:\\
<info> \\}
\end{tcolorbox}

\begin{tcolorbox}[
    breakable,
    title=Prompt Templates of Visual Responder,
    colback=white,               
    colframe=blue!50!black,        
    fonttitle=\bfseries\itshape,   
    arc=2pt,               
    left=5mm,                 
    right=5mm,
    top=3mm,
    bottom=3mm
]
\texttt{Please answer my question in a tone that provides a concise description of the image. If it is a yes/no question, focus on describing the relevant visual information, avoiding answering with yes/no.\\
\\
Question:\\
<question>\\}
\end{tcolorbox}

\begin{tcolorbox}[
    breakable,
    title=Prompt Templates of Summarizer,
    colback=white,           
    colframe=blue!50!black,      
    fonttitle=\bfseries\itshape,  
    arc=2pt,                     
    left=5mm,                    
    right=5mm,
    top=3mm,
    bottom=3mm
]
\texttt{The following is the available information: \\
<info>\\
\\
Please solve the following problems step by step: \\
<question>\\
\\
Use the following format: \\
Thought: Conduct an analysis before you give me an answer.\\
Final Answer: "The final answer you get when you have finished reasoning."\\}
\end{tcolorbox}

\begin{tcolorbox}[
    breakable,
    title=Prompt Templates of Cohesion Enhancement,
    colback=white,                 
    colframe=blue!50!black,        
    fonttitle=\bfseries\itshape,   
    arc=2pt,                      
    left=5mm,                     
    right=5mm,
    top=3mm,
    bottom=3mm
]
\texttt{Below is the reasoning steps for the question <Question>, but there are some disjointed parts marked with "...". Please fill in the gaps to improve coherence. You can use some connecting phrases such as "Let's double check," "Let's check the image again," and "To sum up," and "Wait". \\
\\
Use the following format: \\
{'Thought': 'Reasoning steps', 
 'Final answer':'$\backslash$boxed\{...\}'}
\\
The final answer (only choice like A, B, C, D) MUST BE put in $\backslash$boxed\{\}.\\ 
\\
The reasoning steps is:\\
""" \\
<Reasoning>\\
"""\\}
\end{tcolorbox}

\begin{tcolorbox}[
    breakable,
    title=Prompt Templates of RL Training and Evaluation,
    colback=white,            
    colframe=blue!50!black,       
    fonttitle=\bfseries\itshape,    
    arc=2pt,                      
    left=5mm,                       
    right=5mm,
    top=3mm,
    bottom=3mm
]
\texttt{You FIRST think about the reasoning process as an internal monologue
and then provide the final answer.\\
The reasoning process MUST BE enclosed within <think> </think> tags. The final answer MUST BE put in $\backslash$boxed\{\}.\\
Qustion:}
\end{tcolorbox}
\section{Data Resources}
\label{apppenidx:data_statistic}

We collect data from a large multimodal corpus for (1)  constructing reasoning data with visual reflection pattern (cold-start initialization stage) and (2) GRPO training, as summarized in Tables X and Y.

\begin{table}[t]
\caption{Detailed composition of the datasets used to construct reasoning data with visual-reflection pattern for cold-start initialization.}
\centering

\begin{tabular}{@{}lc@{}}
\toprule
Datasets & Samples\\
\midrule
AI2D \cite{kembhavi2016diagram} &  $\sim$ 0.5K\\
A-OKVQA \cite{marino2019ok} & $\sim$ 0.5K\\
M3CoT (train set) \cite{chen2024m3cot} & $\sim$ 1.0K\\
CLEVR-Math \cite{johnson2017clevr} & $\sim$ 0.5K\\
ScienceQA \cite{masry2022chartqa} & $\sim$ 0.5K\\
TextVQA \cite{singh2019towards} & $\sim$ 0.2K\\
\bottomrule
\end{tabular}

\label{tab:supp-data}
\end{table}

\begin{table}[t]
\caption{Detailed composition of the datasets used to conduct GRPO.}
\centering

\begin{tabular}{@{}lc@{}}
\toprule
Datasets & Samples\\
\midrule
Geo3K \cite{lu2021inter} & $\sim$ 2.1K\\
AI2D \cite{kembhavi2016diagram} & $\sim$ 1.5K\\
TextVQA \cite{singh2019towards}  & $\sim$ 0.8K\\
M3CoT (train set) \cite{chen2024m3cot} & $\sim$ 3.0K\\
MathVerse \cite{zhang2024mathverse} & $\sim$ 2.5K\\
Super-CLEVR \cite{li2023super} & $\sim$ 0.5K\\
MathV360K \cite{shi2024math} & $\sim$ 1.0K\\
A-OKVQA \cite{marino2019ok} & $\sim$ 0.5K\\
ScienceQA  \cite{schwenk2022okvqa}   & $\sim$ 0.5K\\
ChartQA \cite{masry2022chartqa} & $\sim$ 1.0K\\
ArxivQA \cite{li2024multimodal} & $\sim$ 1.0K\\
EMMA \cite{hao2025can} & $\sim$ 1.6K\\
\bottomrule
\end{tabular}

\label{tab:reflection-data}
\end{table}

\section{Supplementary Experiments}
\label{apppenidx:sup_exp}

\begin{figure*}[t]
  \centering
  \includegraphics[width=0.999\textwidth]{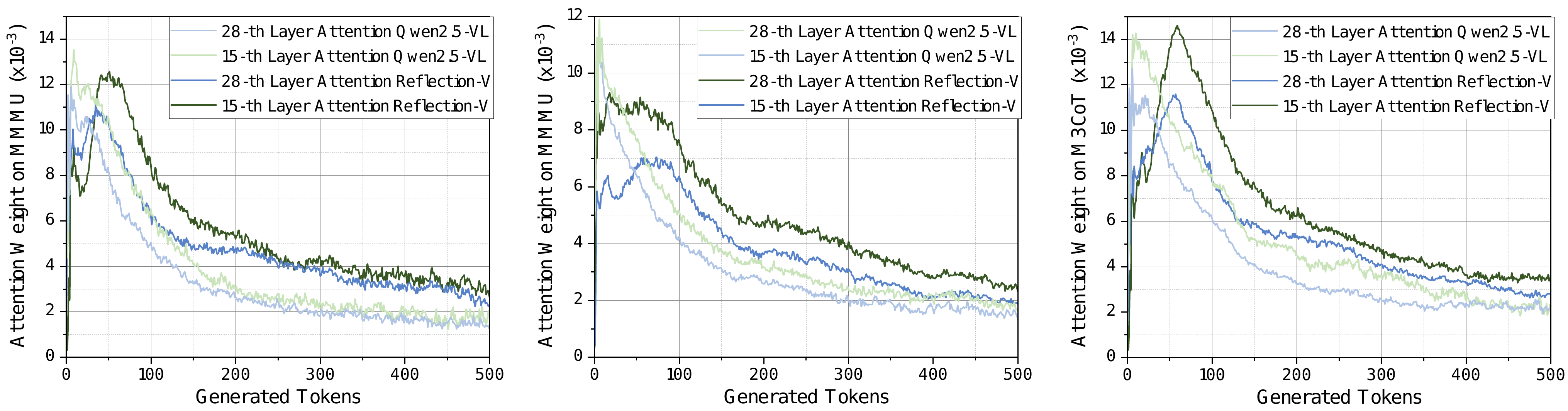}
  \caption{Attention weights (last layer) on visual tokens of Reflection-V-7B on multiple benchmarks, compared to OpenVLThinker-7B and Qwen2.5VL-7B. The shown attention weights represent the mean value across all samples. 
  }
  \label{fig:atten_all}
\end{figure*}

\begin{figure*}[t]
  \centering
  \includegraphics[width=0.999\textwidth]{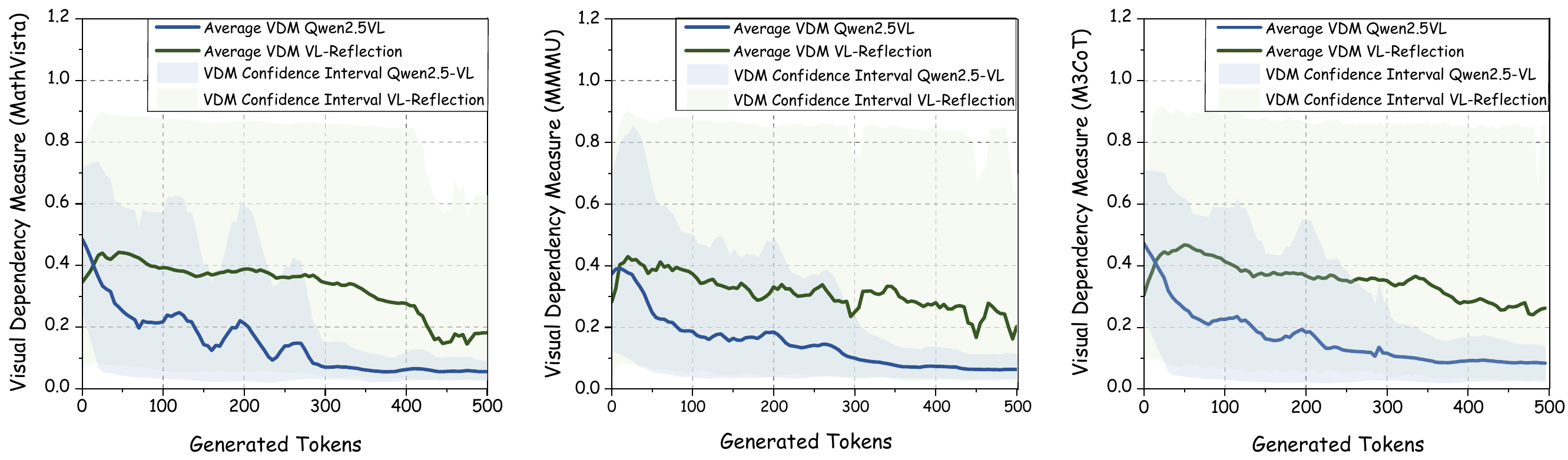}
  \caption{Visual dependency measur of Reflection-V-7B on multiple benchmarks, compared to OpenVLThinker-7B and Qwen2.5VL-7B. Visual dependency measure quantifies the difference in probability distributions for next token prediction based on generated tokens, before and after discarding visual tokens.
  }
  \label{fig:vdm_all}
\end{figure*}

\textbf{Broader evaluation of Reflection-V’s capability of sustained visual attention.} Beyond the MMMU results reported in the main text, Figures 6 and 7 evaluate Reflection-V-7B on three additional visual-reasoning benchmarks spanning mathematical, multi-disciplinary, and general domains. Figure 6 traces last-layer attention from response tokens to visual tokens over 500 generated tokens: Reflection-V-7B consistently launches with higher visual-attention strength than Qwen2.5VL-7B and—crucially—decays far more slowly, retaining about 30\%–40\% of its initial level where baselines sink below 10\%. This advantage extends to other VRMs such as OpenVLThinker-7B, whose cold-start and RL stages, as noted earlier, further erode visual attention and dependency; Reflection-V therefore surpasses these models as well. Figure 7 reports the Visual Dependency Measure (VDM): the upper bound of Reflection-V’s confidence interval remains nearly flat across all tasks, indicating sustained reliance on visual evidence, while the baselines exhibit a pronounced downward trend. These supplementary findings demonstrate that Reflection-V maintains robust visual attention and dependency across diverse reasoning scenarios, substantiating its superior visual-reflection capability.

\textbf{Scaling experiment of the proposed method.} We further conduct a scaling experiment on the MS-SWIFT framework, leveraging LoRA as an efficient training technique on InternVL3-14B. The number of GRPO training epochs is also set to 12. The evaluation results on several visual reasoning benchmarks are shown in Table A. These results demonstrate that our method significantly outperforms GRPO, which solely relies on textual output-based reward. This experiment demonstrates that our approach can effectively scale to larger models and enhance their visual reasoning capabilities.

\begin{table*}[tb]
\small
\centering
\setlength{\tabcolsep}{4pt}  
\renewcommand{\arraystretch}{1.25} 
\begin{tabular}{lcccccc}
\toprule
\textbf{Model} & \multicolumn{2}{c}{\textbf{Math-Reasoning}} & \multicolumn{2}{c}{\textbf{Multi-Disciplinary}} & \multicolumn{1}{c}{\textbf{General}}   \\
\cmidrule(lr){2-3} \cmidrule(lr){4-5} \cmidrule(lr){6-6} 
 & MathVision & MathVista & MMMU & MMMU-Pro & M3CoT \\
\midrule
\multicolumn{6}{c}{\textbf{Closed-Source Vision-Language Models}} \\
\midrule
GPT-4o \cite{hurst2024gpt} & 30.4 & 60.0 & 69.1 & 51.9 & $^\dagger$74.2 \\
GPT-4V \cite{yang2023dawn} & 22.8 & 49.9 & 56.8 & 41.2 & 62.6 \\
\midrule
\multicolumn{6}{c}{\textbf{Open-Source Vision-Language Models}} \\
\midrule
QwenVL2.5-3B \cite{bai2025qwen2} & 21.2 & 62.3 & $^\dagger$51.2 & 31.6 & $^\dagger$55.6 \\
QwenVL2.5-7B \cite{bai2025qwen2} & 25.1 & 68.2 & $^\dagger$54.3 & 36.9 & $^\dagger$60.5 \\
InternVL-2.5-38B \cite{chen2024expanding} & 32.2 & 71.9 & 57.6 & 46.0 & $^\dagger$68.9 \\
InternVL3-14B \cite{chen2024expanding} & 35.9 & 73.8 & 64.1 & 48.9 & 70.1  \\
\midrule
\multicolumn{6}{c}{\textbf{Open-Source Vision-Language Reasoning Models}} \\
\midrule
MM-Eureka-7B$^\diamond$ \cite{meng2025mm} & 26.9 & 73.0 & $^\dagger$51.3 & $^\dagger$36.7 & $^\dagger$63.5 \\ 
R1-Onevision-7B$^\diamond$ \cite{yang2025r1} & 29.9 & 64.1 & $^\dagger$48.7 & 21.6 & $^\dagger$53.1  \\
OpenVLThinker-7B$^\diamond$ \cite{deng2025openvlthinker} & 25.3 & 70.2 & 52.5 & 37.3 & $^\dagger$62.2  \\
\midrule
\multicolumn{6}{c}{\textbf{Ours} (\textit{Training strategy emphasizing visual reflection})} \\
\midrule
InternVL3-14B (GRPO) & \underline{38.3} & \underline{75.6} & \underline{66.9} & \underline{51.3} & \underline{73.4}   \\
Reflection-V (InternVL3-14B) & \textbf{39.8 }& \textbf{76.8} & \textbf{68.7} & \textbf{53.5} & \textbf{78.1} \\
\bottomrule
\end{tabular}
\caption{\small Performance of the proposed approach when scaled to 14B parameters (InternVL3-14B).}
\label{table:scaling}
\end{table*}

\section{Case Study}
\label{apppenidx:cases}

\begin{figure*}[t]
  \centering
  \includegraphics[width=0.999\textwidth]{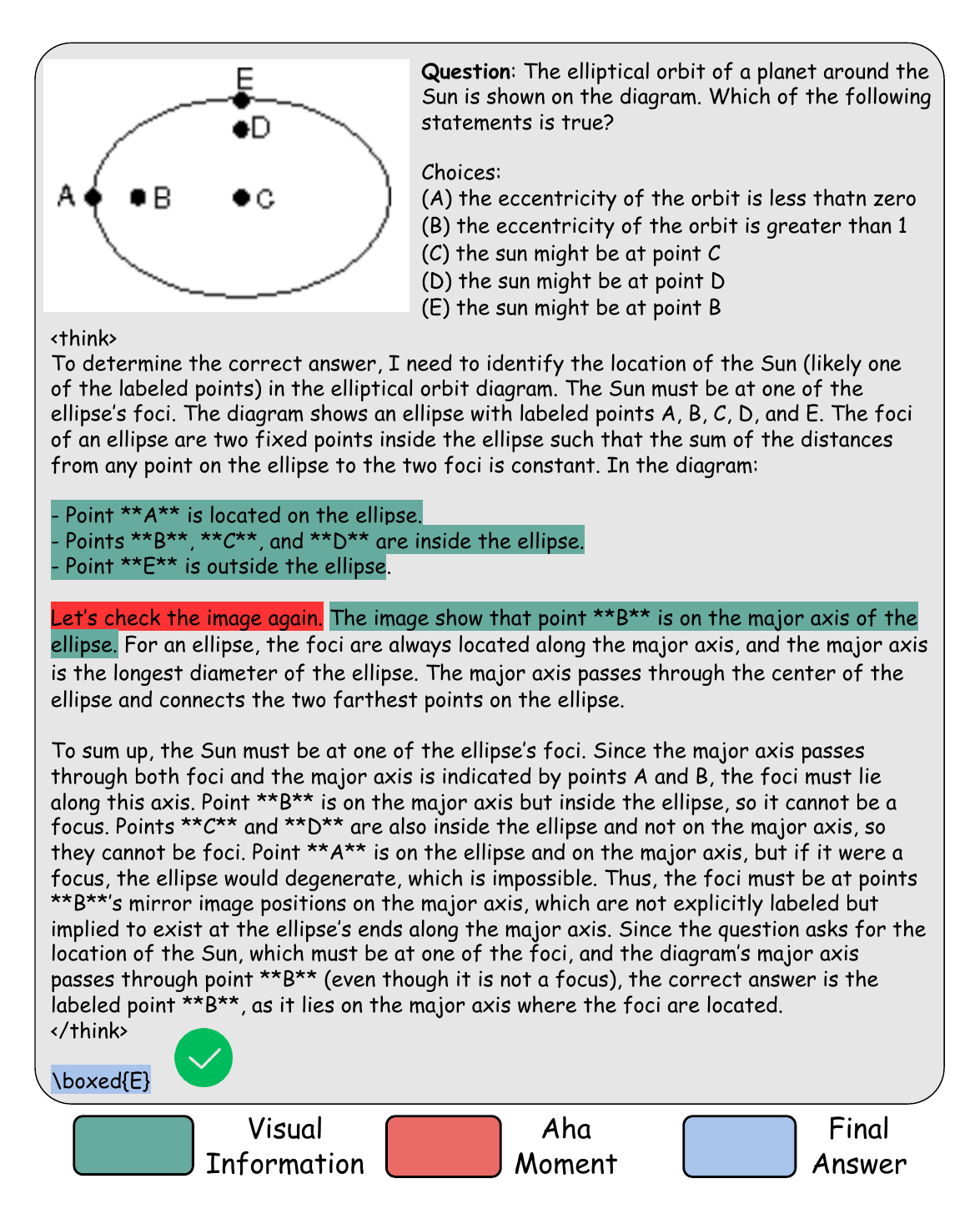}
  \caption{Case study 1 for our model, Reflection-V, performs visual reflection during visual reasoning. In this case, based on rechecking the image, Reflection-V reveals key visual information which is not in the generated reasoning context, therefore finally infers the correct answer.
  }
  \label{fig:case1}
\end{figure*}

\begin{figure*}[t]
  \centering
  \includegraphics[width=0.999\textwidth]{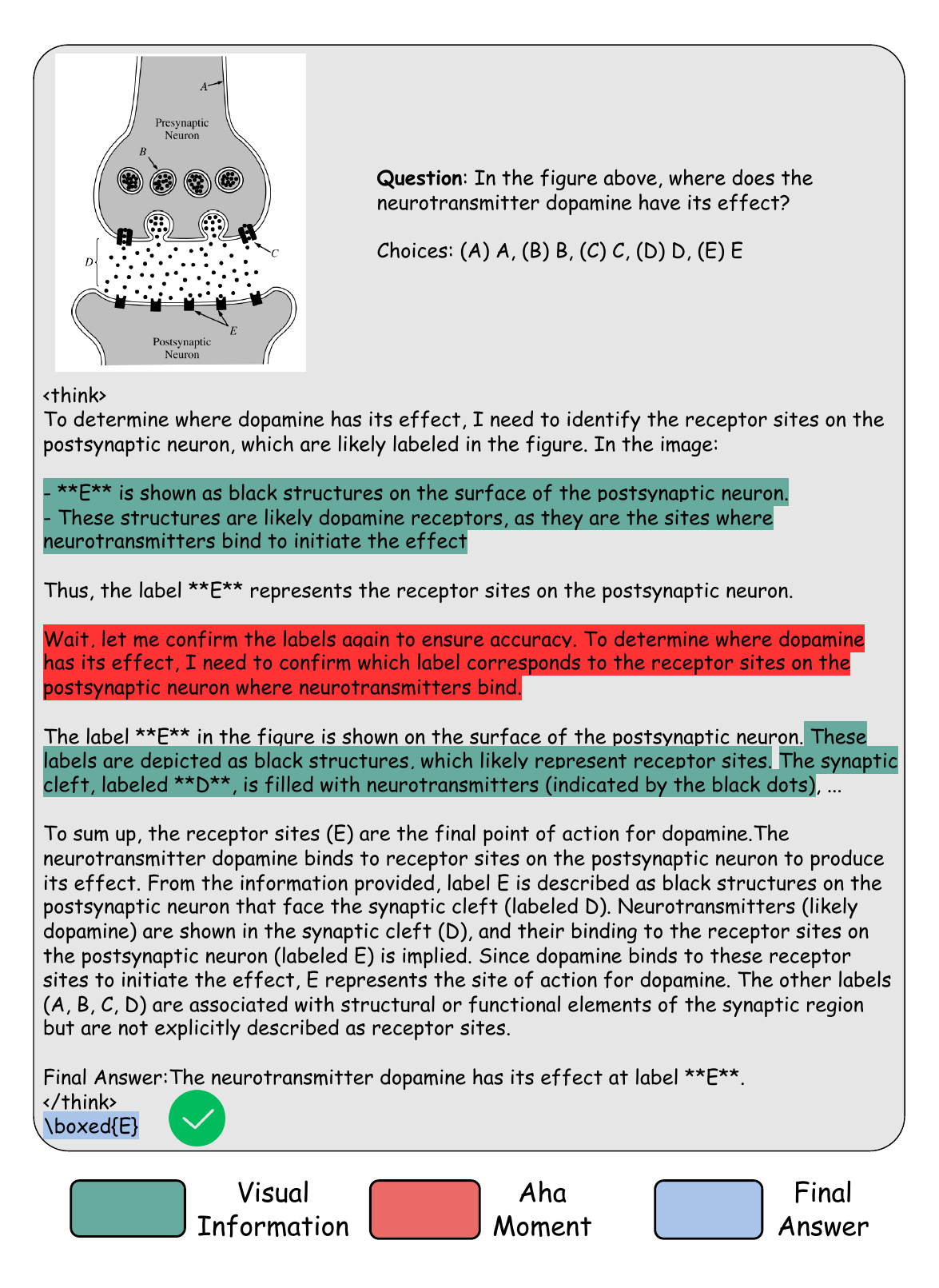}
  \caption{Case study 2 for our model, Reflection-V, performs visual reflection during visual reasoning. In this case, based on rechecking the image, Reflection-V reveals key visual information which is not in the generated reasoning context, therefore finally infers the correct answer.
  }
  \label{fig:case2}
\end{figure*}

\begin{figure*}[t]
  \centering
  \includegraphics[width=0.999\textwidth]{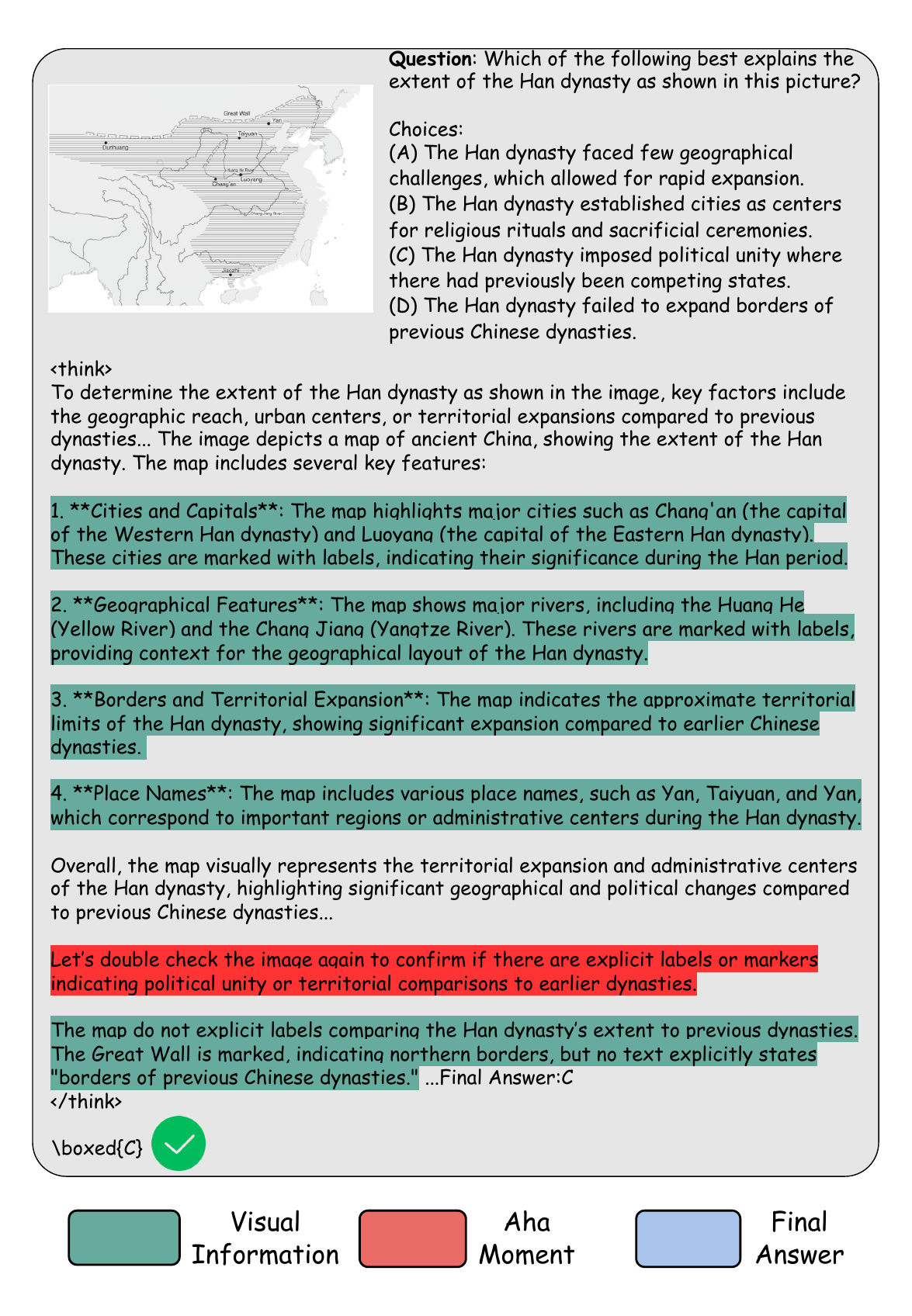}
  \caption{Case study 3 for our model, Reflection-V, performs visual reflection during visual reasoning. In this case, based on rechecking the image, Reflection-V reveals key visual information which is not in the generated reasoning context, therefore finally infers the correct answer.
  }
  \label{fig:case3}
\end{figure*}

\end{document}